\documentclass{article}

\usepackage{arxiv}
\usepackage{array} 
\usepackage[utf8]{inputenc} 
\usepackage[T1]{fontenc}    
\usepackage{hyperref}       
\usepackage{url}            
\usepackage{booktabs}       
\usepackage{amsfonts}       
\usepackage{nicefrac}       
\usepackage{microtype}      
\usepackage{lipsum}
\usepackage{graphicx}
\graphicspath{ {./images/} }
\usepackage{algorithm}
\usepackage{algorithmic}
\usepackage{float}
\usepackage{booktabs}
\usepackage{amsfonts}
\usepackage{bibentry}
\usepackage{amsmath}

\title{Let Experts Feel Uncertainty: A Multi-Expert Label Distribution Approach to Probabilistic Time Series Forecasting}

\author{
 Zhen Zhou \\
  School of Transportation\\
  Southeast University\\
  Nanjing, Jiangsu 213300 \\
  \texttt{zzhou602@seu.edu.cn} \\
   \And
 Zhirui Wang \\
  School of Transportation\\
  Southeast University\\
  Nanjing, Jiangsu 213300 \\
  \texttt{220243664@seu.edu.cn} \\
  \And
 Qi Hong \\
  School of Transportation\\
  Southeast University\\
  Nanjing, Jiangsu 213300 \\
  \texttt{hongqi@seu.edu.cn} \\
  \And
 Yunyang Shi \\
  School of Artificial Intelligence \\and Computer Science\\
  Jiangnan University\\
  Wuxi, Jiangsu 214122 \\
  \texttt{yunyang-shi@seu.edu.cn} \\
  \And
 Ziyuan Gu \\
  School of Transportation\\
  Southeast University\\
  Nanjing, Jiangsu 213300 \\
  \texttt{ziyuangu@seu.edu.cn} \\
    \And
 Zhiyuan Liu \\
  School of Transportation\\
  Southeast University\\
  Nanjing, Jiangsu 213300 \\
  \texttt{zhiyuanl@seu.edu.cn} \\
}

\begin{document}
\bibliographystyle{plain}
\maketitle

\begin{abstract}
Time series forecasting in real-world applications requires both high predictive accuracy and interpretable uncertainty quantification. Traditional point prediction methods often fail to capture the inherent uncertainty in time series data, while existing probabilistic approaches struggle to balance computational efficiency with interpretability. We propose a novel Multi-Expert Learning Distributional Labels (LDL) framework that addresses these challenges through mixture-of-experts architectures with distributional learning capabilities. Our approach introduces two complementary methods: (1) Multi-Expert LDL, which employs multiple experts with different learned parameters to capture diverse temporal patterns, and (2) Pattern-Aware LDL-MoE, which explicitly decomposes time series into interpretable components (trend, seasonality, changepoints, volatility) through specialized sub-experts. Both frameworks extend traditional point prediction to distributional learning, enabling rich uncertainty quantification through Maximum Mean Discrepancy (MMD). We evaluate our methods on aggregated sales data derived from the M5 dataset, demonstrating superior performance compared to baseline approaches. The continuous Multi-Expert LDL achieves the best overall performance, while the Pattern-Aware LDL-MoE provides enhanced interpretability through component-wise analysis. Our frameworks successfully balance predictive accuracy with interpretability, making them suitable for real-world forecasting applications where both performance and actionable insights are crucial.
\end{abstract}


\section{Introduction}
Time series forecasting is a cornerstone of decision-making in fields ranging from finance to healthcare, yet it remains fundamentally challenging due to the complex, dynamic nature of real-world data \cite{eldele2024label}. Traditional models often fall short by producing rigid, one-size-fits-all predictions—either as oversimplified point estimates or as probabilistic outputs constrained by restrictive parametric assumptions. These limitations become glaringly apparent in scenarios where data exhibits heterogeneous patterns and complex uncertainty. For instance, in retail forecasting, promotional events may create sudden demand spikes with disconnected possible outcomes (e.g., a 50\% chance of moderate success vs. 50\% chance of viral demand), defying the smooth, unimodal distributions assumed by conventional probabilistic models. Beyond quantifying uncertainty, proper distributional modeling actively enhances prediction accuracy by preventing overconfidence in ambiguous regimes and enabling pattern-specific error correction—whether adapting to volatile market shocks or refining trend estimates through distribution-aware learning. This dual capability emerges because representing the full predictive distribution provides richer training signals and more robust optimization landscapes compared to point estimation alone.

The forecasting community has long approached these challenges in isolation. On one hand, Mixture of Experts (MoE) \cite{gan2025mixture} architectures excel at decomposing temporal complexity by routing different patterns to specialized submodels\cite{han2022dynamic}. Yet, they typically output point estimates, ignoring uncertainty altogether—a critical flaw when decisions require risk quantification (e.g., inventory planning for products with intermittent demand). On the other hand, Label Distribution Learning (LDL) \cite{geng2016label} first time offers a flexible, non-parametric way to model arbitrary uncertainty structures—but when applied to time series, it often relies on monolithic architectures incapable of handling diverse temporal regimes. This disconnect forces practitioners to choose between modeling patterns well or capturing uncertainty accurately, leaving them ill-equipped for scenarios where both the shape and magnitude of uncertainty vary by pattern type.

We address these limitations through Multi-Expert LDL and its advanced extension, Pattern-Aware LDL-MoE—a novel framework that unifies architectural specialization with flexible distributional modeling for time series forecasting. The foundation of our approach lies in a crucial observation: uncertainty in temporal data is fundamentally regime-dependent and pattern-specific. For example, in supply chain forecasting, stable inventory demand typically follows a low-variance normal distribution (±5\% fluctuation), while pandemic disruptions create a bi-modal uncertainty pattern—with 60\% probability of minor delays (1-2 weeks) and 40\% risk of severe shortages (8+ weeks)—demanding fundamentally different modeling approaches for each regime \cite{zhou2024urban}. Our framework captures this complexity by dynamically pairing each pattern type (trend, seasonality, changepoints, volatility) with specialized distributional heads, enabling precise adaptation to diverse uncertainty regimes. Beyond accuracy, the model provides interpretable uncertainty attribution, clearly identifying whether prediction uncertainty stems from measurement noise (volatility expert), periodic variations (seasonal expert), or anomalous events (changepoint expert). This capability represents a paradigm shift—from conventional forecasting as passive prediction to decision-centric risk intelligence, where uncertainty explanations drive actionable business strategies.

\section{Related Work}

The foundations of probabilistic time series forecasting rest on three research strands that have evolved both independently and in parallel. Tracing their development reveals the critical gaps our work addresses.

The quest to quantify predictive uncertainty began with Bayesian approaches, where MC-Dropout \cite{gal2016dropout} and Deep Ensembles \cite{lakshminarayanan2017simple} treated model parameters as distributions rather than point estimates. While seminal, these methods face fundamental limitations in temporal domains—MC-Dropout's stochastic forward passes introduce prohibitive latency for real-time forecasting, and both approaches conflate epistemic and aleatoric uncertainty. Quantile regression \cite{wen2017multi} emerged as a computationally efficient alternative, directly modeling prediction intervals through percentiles. However, its piecewise optimization often produces incoherent distributions (quantile crossings) and fails to capture inter-quantile dependencies critical for scenario analysis. These limitations spurred the development of LDL \cite{geng2016label,gu2025topological}, which treats the entire output distribution as a learnable target. LDL's non-parametric discretization handles multi-modality naturally—a crucial advantage for time series where regime shifts create disjoint outcome possibilities. However, LDL integration with modern temporal architectures remains superficial, often treating time series as independent windows rather than evolving processes.

Parallel advances in model architecture addressed temporal heterogeneity through increasingly sophisticated designs. The MoE framework \cite{jacobs1991adaptive} pioneered conditional computation, where specialized submodels handle distinct input patterns. Modern sparse gating implementations \cite{shazeer2017outrageously} scaled this concept to hundreds of experts while maintaining computational efficiency through intelligent routing. Time series adaptations \cite{lai2018modeling, wu2020connecting} demonstrated remarkable accuracy gains by combining MoE with temporal modules—LSTMs for sequential dependencies, CNNs for local patterns, and attention mechanisms for long-range interactions. Yet these advances focused exclusively on point prediction, creating an ironic divergence: while MoEs became exceptionally skilled at identifying and processing different temporal regimes (trends, seasonality, shocks), they provided no mechanism to characterize the distinct uncertainty profiles of these regimes. This left practitioners with sharp point forecasts but no way to assess their reliability across different temporal contexts.

The probabilistic forecasting literature developed its own architectural conventions, often constrained by parametric assumptions. DeepAR \cite{salinas2020deepar} and related autoregressive models demonstrated that deep networks could learn complex temporal dynamics while outputting Gaussian or negative binomial distributions. Subsequent innovations like Gaussian Copula processes \cite{salinas2019high} added flexibility in modeling dependencies across time steps and series. However, these approaches share a critical limitation: their parametric output distributions cannot represent the multi-modal, regime-dependent uncertainties prevalent in real-world systems. Even state-of-the-art CRPS-based methods \cite{salinas2019high}, while achieving excellent calibration scores, provide limited insight into the structural sources of uncertainty—whether a prediction's variance stems from measurement noise, regime ambiguity, or model uncertainty.

Our work bridges these research trajectories through three key unifications: (1) We integrate the architectural flexibility of MoE with the distributional expressiveness of LDL, empowering experts to specialize in both pattern recognition and uncertainty characterization. (2) We develop a time series decomposition technique that propagates uncertainty information across related time points and series, creating more informative learning targets; (3) We introduce a gating regularization framework that maintains the benefits of expert specialization while preventing collapse. This synthesis moves beyond the artificial dichotomy between architectural complexity and distributional flexibility—instead showing how specialized architectures can directly enable more nuanced uncertainty quantification. The result advances both theoretical foundations and practical outcomes.

\section{Methodology}
\subsection{Framework Overview}
\begin{figure*}[t]
    \centering
    \includegraphics[width=\textwidth]{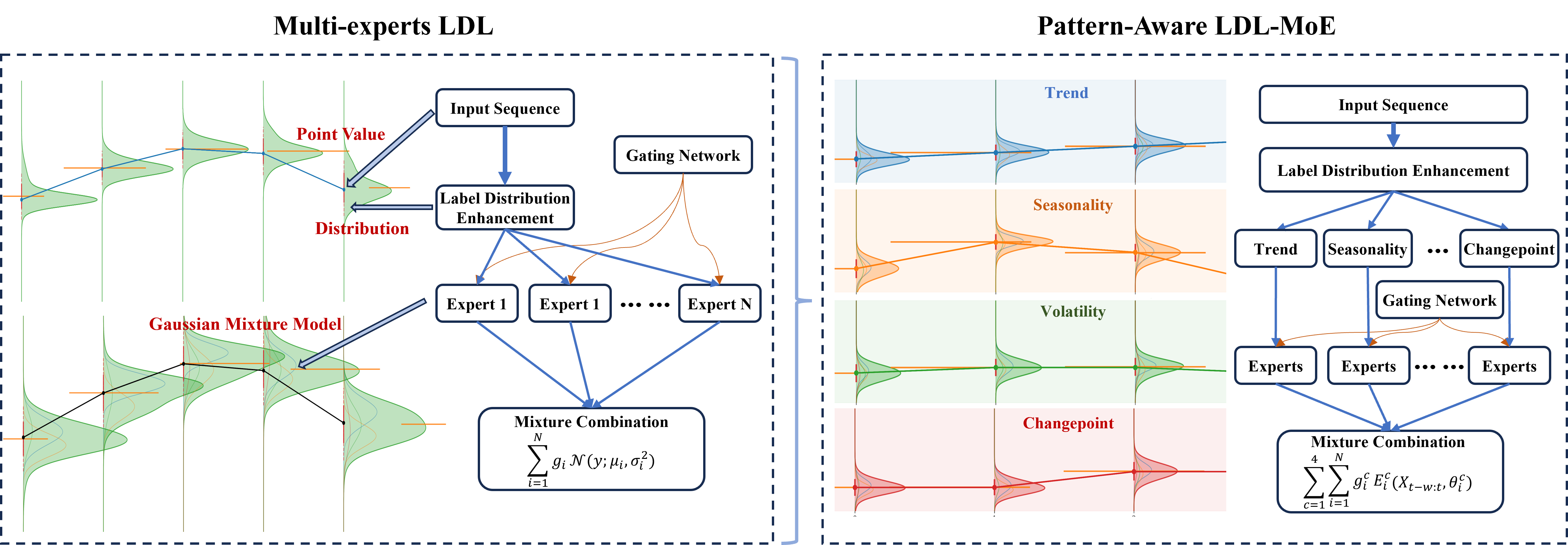}
    \caption{The framework of Multi-experts LDL and Pattern-Aware LDL-MoE}
    \label{fig1}
\end{figure*}

We propose two complementary pipelines for probabilistic forecasting, shown in Figure~\ref{fig1}. The primary architecture is the \textbf{Multi-Expert LDL}, which uses discrete or continuous label distributions and MoE. The second is the \textbf{Pattern-aware LDL-MoE}, which combines STL-style decomposition with pattern recognition and uncertainty quantification. It's designed to automatically identify and model different types of time series patterns (trend, seasonal, changepoint, volatility) while providing probabilistic forecasts. The Multi-Expert LDL and Pattern-aware LDL-MoE all have the enhanced label distribution enhancement, but they are different in  multi-expert processing, gating and combination, and a composite loss function.

\subsection{Label Distribution Enhancement}
The quality of a learned model is bounded by the quality of its learning target. For LDL, this means the target distribution must be as informative as possible. We move beyond simple, static assumptions to create a target distribution whose shape is inferred directly from the underlying structure of the data. This data-driven approach allows the model to learn from a more nuanced representation of uncertainty. We convert point labels \( y \) from a batch of size \( B \) into structured probability distributions through a three-step process:

\begin{enumerate}
    \item \textbf{Similarity Analysis}: The insight here is that time series exhibiting similar patterns are likely to share similar uncertainty profiles. This is particularly valuable in data-sparse contexts where a series can "borrow" statistical strength from its neighbors. We find the KNN for each time series \( x_i \) in the input batch using a KD-Tree and weight their relationship using a Gaussian kernel:
    \[
    w_{ij} = \exp\left(-\frac{\|x_i - x_j\|^2}{2\sigma^2}\right)
    \]
    
    \item \textbf{Periodicity Detection}: Uncertainty is often not constant but correlated with cyclical patterns (e.g., higher volatility during peak business hours). We identify the dominant period \( p \) for each series by finding the lag \( \tau \) that maximizes the autocorrelation function, thus capturing cyclical variations in uncertainty:
    \[
    \rho(\tau) = \frac{\sum_{t=1}^T (x_t - \bar{x})(x_{t-\tau} - \bar{x})}{\sum_{t=1}^T (x_t - \bar{x})^2}
    \]
    
    \item \textbf{Variance Smoothing}: A naive variance estimate can be noisy. We use graph regularization to enforce smoothness and consistency. We first compute a base variance \( \sigma_{\text{base}} \) using a sliding window. Then, we construct a graph where nodes are time steps and edges encode temporal adjacency, periodic links, and cross-series similarity. The graph Laplacian acts as a smoothing filter, propagating variance information across related time points. The final smoothed variance \( \sigma \) is obtained by solving this graph-regularized linear system:
    \[
    \sigma = (I + \lambda L)^{-1} v_{\text{base}}
    \]
    where \( L = D - A \) is the graph Laplacian \cite{xu2019label} built from the weighted adjacency matrix \( A \), which is obtained from similarity analysis and periodicity detection, and \( \lambda \) is a regularization parameter. This smoothed variance is then used to generate the final target distribution.
\end{enumerate}

\subsection{Multi-Expert LDL}

The ``no free lunch'' theorem suggests that no single model architecture is universally optimal for all types of data patterns. A time series is often a composite of multiple underlying processes. Our multi-expert architecture embraces this reality by creating an ensemble of specialists, each with the same architecture but different learned parameters, allowing them to specialize in different aspects of the time series through training. Multiple LSTM experts process the input sequence $X \in \mathbb{R}^{T \times d}$. Each expert is implemented as a bidirectional LSTM with Gaussian output heads. The recurrent nature and memory cells are ideal for modeling phenomena where the order of events is critical, such as autoregressive trends and sequential dependencies.
    \[
    h_t = \mathrm{LSTM}(x_t, h_{t-1})
    \]
    Each expert outputs both mean and variance parameters:
    \[
    (\mu_i, \log\sigma_i^2) = \mathrm{MLP}_{\text{mean}}(h_T) \oplus \mathrm{MLP}_{\text{var}}(h_T)
    \]

\subsubsection{Gating \& Combination}
With a team of specialists, we need an intelligent orchestrator. The gating network serves this role, learning to dynamically analyze the input data and route it to the most qualified expert(s), effectively performing a learned model selection for each individual forecast. A gating network, implemented as an MLP, computes a weight vector $g$ for the experts. We use a temperature-controlled softmax function, where the temperature $\tau$ controls the sharpness of the gating decision. A lower temperature leads to a more decisive, sparse selection (exploitation), while a higher temperature results in a softer, more distributed allocation (exploration):
\[
g = \mathrm{softmax}(\tau^{-1} \cdot \mathrm{MLP}(x)), \quad \tau > 0
\]
For window size $w$, the final predicted distribution $\hat{y}$ is the weighted sum of the individual expert outputs, allowing for a flexible combination of their specialized views:
\[
\hat{y}_t = \sum_{i=1}^N g_i \cdot \mathrm{E}_i(X_{t-w:t})
\]

\subsubsection{Distributional Learning Framework}
Our multi-expert LDL framework extends traditional point prediction to distributional learning, where each expert learns to predict the full distribution of the target variable rather than just its expected value. This approach accommodates both discrete and continuous target distributions.

For categorical or discrete targets, each expert outputs parameters of a categorical distribution:
\begin{align*}
\mathbf{z}_i &= E_i(X, \theta_i) \\
p_i(y = k) &= \frac{\exp(z_{i,k})}{\sum_{j=1}^K \exp(z_{i,j})} \\
p_{\mathrm{mixture}}(y = k) &= \sum_{i=1}^N g_i \cdot p_i(y = k)
\end{align*}

For continuous targets, each expert outputs Gaussian parameters (mean and variance), and the final output is a mixture gaussian distribution:
\begin{equation}
\begin{split}
(\mu_i, \log\sigma_i^2) &= E_i(X, \theta_i) \\
p_i(y) &= \mathcal{N}(y; \mu_i, \sigma_i^2) \\
p_{\mathrm{mixture}}(y) &= \sum_{i=1}^N g_i \cdot \mathcal{N}(y; \mu_i, \sigma_i^2)
\end{split}
\end{equation}

Target labels $y$ are mapped to Gaussian parameters with a standard deviation learning from label enhancement. The model is trained by minimizing the distance metrics between the predicted mixture and the target distribution. In the case of discrete distributions, we utilize the Kullback-Leibler (KL) divergence. However, for continuous distributions, due to the absence of a closed form for KL divergence, we primarily rely on MMD (Maximum Mean Discrepancy) \cite{kumagai2019unsupervised} as our distance metric. This choice is motivated by MMD's unique strengths in probabilistic forecasting, which offer distinct advantages for our modeling approach. The MMD between distribution $P$ and $Q$ is
\begin{align*}
\mathrm{MMD}^2(P,Q) = \mathbb{E}_{x,x'\sim P}[k(x,x')] +\\ \mathbb{E}_{y,y'\sim Q}[k(y,y')] - 2\mathbb{E}_{x\sim P,y\sim Q}[k(x,y)]
\end{align*}

For Gaussian mixtures, MMD permits exact gradient computation through closed-form kernel evaluations:
\begin{equation}
\mathbb{E}_{x,x'\sim P}[k(x,x')] = \sum_{i,j} g_i g_j \frac{ \kappa }{\sqrt{\sigma_i^2+\sigma_j^2+ \kappa ^2}}e^{-\frac{(\mu_i-\mu_j)^2}{2(\sigma_i^2+\sigma_j^2+ \kappa ^2)}}
\end{equation}
where $ \kappa $ is bandwidth. We employ RBF kernels due to their closed-form computability with Gaussian mixtures and universal approximation properties, making them particularly suitable for our mixture-of-experts framework.
\begin{equation}
\begin{split}
k(x,y) = \approx 1 - \frac{\|x-y\|^2}{2\sigma^2} + \frac{\|x-y\|^4}{8\sigma^4} - \cdots
\end{split}
\end{equation}
The automatic bandwidth selection via median heuristic:
\begin{equation}
\sigma_t = \mathrm{median}\{\|x_i-x_j\|\}_{i<j}
\end{equation}
makes MMD naturally adapt to different prediction regimes within our MoE framework. We employ Random Fourier Features (RFF) \cite{sriperumbudur2015optimal} approximation to maintain $O(d)$ computational complexity while preserving the theoretical guarantees of MMD.
\begin{equation}
k(x,y) \approx \phi(x)^\top\phi(y), \quad \phi(x) = \sqrt{2/D}\cos(\mathbf{W}x + \mathbf{b})\in \mathbb R^d
\end{equation}
This enables scalable training without sacrificing the metric's ability to discriminate between distributional characteristics.

\subsubsection{Loss Functions}
The model's training objective balance the primary goal of predictive accuracy with crucial regularization needs that ensure the stability and efficiency of the MoE framework. A multi-component loss allows us to optimize for the primary objective (distribution matching) while explicitly enforcing desirable behaviors (expert balance and diversity). The model is trained by minimizing a composite loss function $\mathcal{L}_{\mathrm{total}}$:
\[
\mathcal{L}_{\mathrm{total}} = \mathcal{L}_{\mathrm{distance}} + \alpha \mathcal{L}_{\mathrm{bal}} + \beta \mathcal{L}_{\mathrm{div}}
\]
where:
\begin{itemize}
    \item $\mathcal{L}_{\mathrm{distance}}$ is the distribution distance metric. This is the primary learning signal, pushing the predicted distribution to match the shape of the target distribution.
    \item $\mathcal{L}_{\mathrm{bal}}$ is a load-balancing loss that acts as a ``fairness'' regularizer. It minimizes the variance of the average expert utilization $u_i$ across a batch, discouraging the gating network from neglecting any experts:
    \[
    \mathcal{L}_{\mathrm{bal}}=\mathrm{Var}(u), \quad \text{where} \quad u_i = \frac{1}{B} \sum_{b=1}^B g_{b,i}
    \]
    \item $\mathcal{L}_{\mathrm{div}}$ is a diversity loss that serves as a ``specialization'' regularizer. It penalizes high cosine similarity between the representations $e_i$ from different experts, encouraging them to learn distinct, complementary functions:
    \[
    \mathcal{L}_{\mathrm{div}}=\sum_{i \neq j} \cos(e_i, e_j)
    \]
\end{itemize}
The complete objective combines distribution matching with expert regularization:
\begin{equation}
\mathcal{L} = \mathrm{MMD}^2(P,Q) + \lambda_1\mathrm{Var}(\mathbf{g}) + \lambda_2\sum_{i\neq j}\cos(\mathbf{h}_i,\mathbf{h}_j)
\end{equation}
This unified framework provides a flexible approach to probabilistic forecasting that can handle both discrete and continuous target distributions while maintaining the benefits of multi-expert specialization and robust uncertainty quantification.

The Multi-Expert LDL framework is grounded in a key theoretical insight: the intrinsic connection between MoE architectures and Gaussian mixture models enables universal distribution approximation while maintaining computational tractability. Each expert in our system specializes in distinct temporal regimes, with the gating network dynamically adjusting mixture weights to form a flexible Gaussian mixture distribution. This approach fundamentally generalizes traditional single-Gaussian assumptions, preserving their mathematical advantages (e.g., closed-form likelihood computation) while acquiring the capacity to model complex phenomena including multi-modality, skewness, and heavy-tailed distributions. The resulting architecture achieves an optimal balance between expressiveness and interpretability – combining the theoretical rigor of Gaussian mixtures with the adaptive capability of MoE systems to handle real-world forecasting challenges.

\subsection{Pattern-Aware LDL-MoE}

While standard multi-expert architectures leverage architectural diversity, many real-world time series are best understood as the sum of interpretable components such as trend, seasonality, changepoints, and volatility. The Pattern-Aware LDL-MoE framework models these components by decomposing the forecasting task into additive sub-experts, each specializing in a distinct temporal pattern. This approach enhances interpretability, robustness, and the ability to capture complex temporal dynamics.

\subsubsection{Additive Decomposition with Sub-Experts}

Given an input sequence $X \in \mathbb{R}^{T \times d}$, we propose a decomposition-based forecasting framework that extends the Prophet model \cite{taylor2018forecasting} through four specialized experts modeling trend, seasonality, changepoints, and volatility patterns. We introduce a dedicated volatility expert to capture complex non-linear dependencies. Each expert $E_{i}$ follows a MoE architecture, consisting of multiple sub-experts that specialize in distinct temporal regimes: 
\begin{align*}
\text{Trend}_t &= \sum_i^N g_i E_{i}^{\text{trend}}(X_{t-w:t}, \theta_i^{\text{trend}}) \\
\text{Seasonal}_t &= \sum_i^N g_i E_{i}^{\text{seasonal}}(X_{t-w:t}, \theta_i^{\text{seasonal}}) \\
\text{Changepoint}_t &= \sum_i^N g_i E_{i}^{\text{changepoint}}(X_{t-w:t}, \theta_i^{\text{changepoint}}) \\
\text{Volatility}_t &= \sum_i^N g_i E_{i}^{\text{volatility}}(X_{t-w:t}, \theta_i^{\text{volatility}})
\end{align*}

The output is the sum of its expert predictions, and the weighted sum of sub-experts:
\[
\hat y_t = \text{Trend}_t + \text{Seasonal}_t + \text{Changepoint}_t + \text{Volatility}_t
\]

\subsubsection{Distributional Learning and Loss Functions}

Each sub-expert can be designed to output either discrete or continuous distributions, enabling the model to perform LDL at the component level. The overall predictive distribution is thus a mixture of component-wise distributions, aggregated through the gating mechanism.

The training objective combines several terms:
\begin{itemize}
    \item \textbf{Distribution Matching Loss:} For each component, a distance metric (e.g., KL divergence for discrete, MMD for continuous) is used to match the predicted and target distributions.
    \item \textbf{Component-Specific Regularization:} Each sub-expert may have additional regularization, such as smoothness for trend, periodicity for seasonality, sparsity for changepoints, and heteroscedasticity for volatility.
    \item \textbf{Expert Balance and Diversity:} As in standard MoE, load-balancing and diversity losses are included to ensure all experts contribute and learn distinct functions.
    
\end{itemize}

The total loss is:
\[
\mathcal{L}_{\text{total}} = \mathcal{L}_{\text{dist}} + \alpha \mathcal{L}_{\text{bal}} + \beta \mathcal{L}_{\text{div}} + \sum_{c} \lambda_c \mathcal{L}_{\text{reg}}^{(c)}
\]
where $\mathcal{L}_{\text{dist}}$ is the sum of distributional losses across all components, $\mathcal{L}_{\text{bal}}$ and $\mathcal{L}_{\text{div}}$ are as previously defined, and $\mathcal{L}_{\text{reg}}^{(c)}$ are component-specific regularization terms. The component-specific regularization term ensure each sub-expert in the Pattern-Aware LDL-MoE framework employs specialized regularization terms to enforce appropriate behavior for their designated temporal patterns. The trend sub-expert encourages smooth, persistent patterns:
$$\lambda_{\text{smooth}} \|\nabla^2 \text{Trend}_t\|_2^2 + \lambda_{\text{persist}} \|\nabla \text{Trend}_t\|_2^2$$
where the smoothness term penalizes sudden changes and the persistence term controls trend evolution rate. The seasonal sub-expert maintains periodicity and smooth patterns:
$$\lambda_{\text{period}} \|\text{Seasonal}_t - \text{Seasonal}_{t-p}\|_2^2 + \lambda_{\text{smooth}} \|\nabla \text{Seasonal}_t\|_2^2$$
where the periodicity term enforces consistency with seasonal period $p$ and the smoothness term encourages gradual transitions. The changepoint sub-expert detects sparse, localized structural breaks:
$$\lambda_{\text{sparse}} \|\text{Changepoint}_t\|_1 + \lambda_{\text{local}} \|\nabla \text{Changepoint}_t\|_2^2$$
where the sparsity term encourages sparse detection and the localization term ensures focused changepoints. The volatility sub-expert models heteroscedasticity and time-varying uncertainty:
$$\lambda_{\text{hetero}} \|\text{Volatility}_t - \text{Var}(y_t)\|_2^2 + \lambda_{\text{smooth}} \|\nabla \text{Volatility}_t\|_2^2$$
where the heteroscedasticity term aligns with empirical variance and the smoothness term ensures gradual volatility evolution. These regularization terms ensure each sub-expert learns appropriate representations for its designated temporal component while maintaining model coherence and interpretability.

This decomposition enables practitioners to attribute forecast uncertainty to specific sources (e.g., increased volatility or a detected changepoint), greatly enhancing interpretability and actionable insight. By explicitly modeling and aggregating interpretable temporal patterns within a distributional mixture-of-experts framework, Pattern-Aware LDL-MoE achieves both high predictive performance and transparent, component-wise uncertainty quantification.

\subsection{Expert Collapse Mitigation}
Expert collapse is the Achilles' heel of MoE models \cite{singh2024exposing}. Without explicit intervention, training dynamics can lead to a state where only one or two experts ever get selected, rendering the multi-expert architecture useless. Our mitigation strategy is a three-pronged, synergistic approach to ensure robust and stable training.

\begin{enumerate}
    \item \textbf{Temperature Scaling}: We set \( \tau = 1.5 \) in the gating network's softmax. This softens the probability distribution over experts, encouraging exploration and preventing the gating network from becoming overly confident in a single expert too early in training. It acts as a baseline level of exploration.
    
    \item \textbf{Load Balancing Loss}: The explicit loss term \( \mathcal{L}_{\text{bal}} \) directly penalizes unbalanced expert usage. This provides a deterministic gradient signal that pushes the gating network towards more uniform expert selection, complementing the stochastic approaches.
    
    \item \textbf{Noise Injection}: During training, we add small Gaussian noise to the gating network's logits before the softmax activation. This stochastic perturbation helps the model escape poor local minima where one expert is dominant, forcing it to re-evaluate its choices:
    \[
    g_{\text{logits}} \leftarrow g_{\text{logits}} + \epsilon, \quad \epsilon \sim \mathcal{N}(0, 0.1^2).
    \]
\end{enumerate}



\section{Experiments}

\begin{table*}[t]
\centering
\caption{Experiments on Different Models}
\label{tab:exp}·
\begin{tabular}{>{\centering\arraybackslash}p{0.4\linewidth}>{\centering\arraybackslash}ccc>{\centering\arraybackslash}p{0.2\linewidth}@{}}
\toprule
\textbf{Model}                               & \textbf{RMSE} & \textbf{MAE} & \textbf{MAPE} \\ \midrule
LSTM                              & 3.418      & 2.971      & 153.938      \\
LightTS                             & 3.504      & 2.795     & 168.717      \\
GRU                                 & 3.563      & 3.158     & 103.085      \\
ITransformer                        & 4.899      & 4.069     & 255.644      \\
DLinear                             & 6.285      & 5.555     & 312.258       \\
Transformer                         & 6.360       & 5.460     & 325.642      \\
Informer                            & 8.745      & 8.050     & 432.304      \\
DeepAR                              & 24.538      & 24.313     & 1079.199      \\
N-BEATS                              & 74.970      & 75.863     & 2974.654       \\ 
\textbf{Multi-experts LDL (continuous)}               & 3.311      & 2.919     & 141.414      \\
\textbf{Pattern-aware LDL-MoE}  & 3.330      & 2.954     & 140.983      \\
\textbf{Multi-experts LDL (discrete)} & 3.362      & 2.884     & 150.277      \\
\bottomrule
\end{tabular}
\end{table*}
\begin{figure*}[t]
    \centering
    \includegraphics[width=\textwidth]{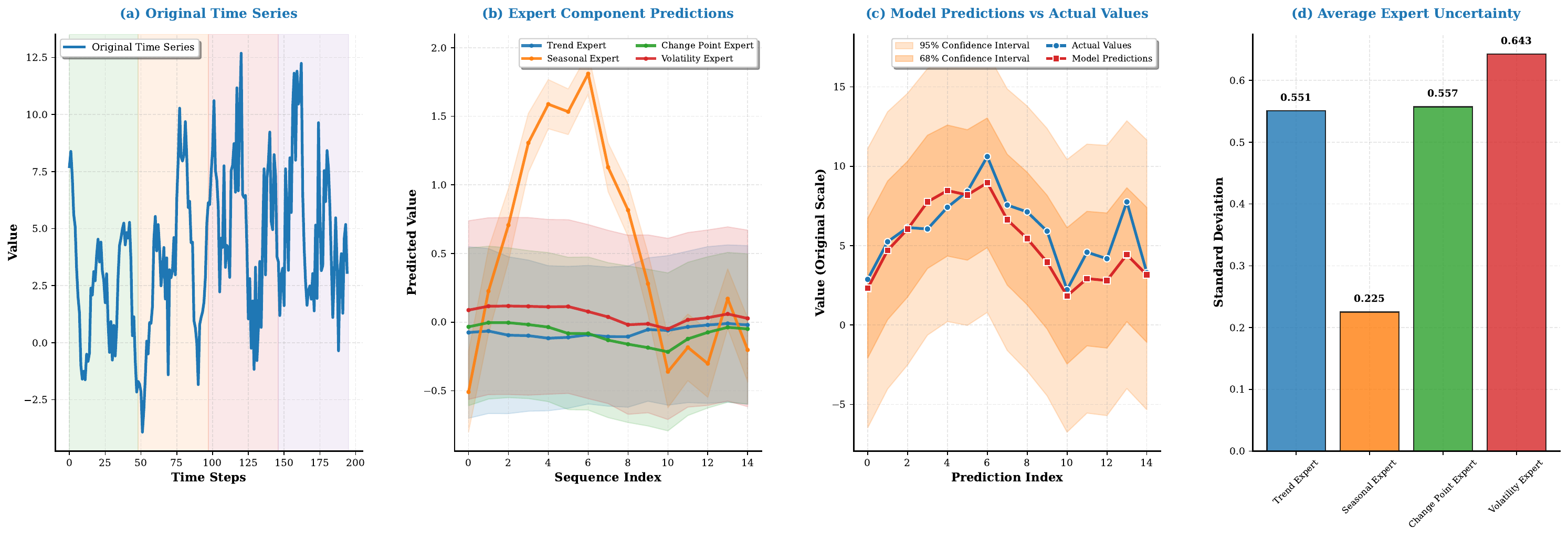}
    \caption{The decomposition and interpretability output of pattern-aware LDL-MoE}
    \label{fig2}
\end{figure*}

\subsection{Dataset and Experimental Setup}
We evaluate our framework on aggregated sales data derived from the M5 Competition dataset \cite{m5-forecasting-accuracy}. Our experimental configuration uses 20-day rolling sequences of sales features (input dimension is 20, sequence length is 20) as inputs, with probabilistic forecasts for the next 28-day period as targets. The dataset's temporal characteristics allow us to rigorously evaluate our model's decomposition capabilities across different time scales. All models are implemented in PyTorch and trained using the Adam optimizer with a learning rate of $1 \times 10^{-3}$. We use early stopping with a patience of 20 epochs based on validation performance to prevent overfitting. For the Multi-Expert LDL frameworks, we configure each model with 4 LSTM experts, each with a hidden dimension of 128 and 2 layers. The gating network uses a temperature parameter $\tau = 1.5$ to balance exploration and exploitation. To ensure realistic evaluation without data leakage, we employ a time-based train/test split where the last 28 days serve as the test set. This approach mimics real-world forecasting scenarios where models must predict future values based only on historical data. The training set is further split into training and validation sets (90\%/10\%) for model selection and hyperparameter tuning.

We compare a diverse set of time series forecasting models, each offering unique strengths for different application scenarios, including DLinear~\cite{zeng2023transformers}, LightTS~\cite{zhang2207less},iTransformer~\cite{liu2023itransformer}, DeepAR~\cite{salinas2020deepar}, N-BEATS~\cite{oreshkin2019n}, Transformer-based models~\cite{vaswani2017attention}, LSTM~\cite{greff2016lstm}, GRU~\cite{dey2017gate}, and Informer~\cite{zhou2021informer}. This spectrum of models highlights the trade-offs between interpretability, computational efficiency, and the ability to provide uncertainty estimates, guiding practitioners in selecting the most appropriate approach for their forecasting needs.

\subsection{Results and Analysis}

Table~\ref{tab:exp} presents the comparative performance of our proposed Multi-Expert LDL frameworks. The results demonstrate the effectiveness of our approach in probabilistic time series forecasting.

\textbf{Continuous vs. Discrete Distribution Modeling} The continuous Multi-Experts LDL approach achieves the best overall performance with an RMSE of 3.311, MAE of 2.919, and MAPE of 141.41. This superior performance can be attributed to several advantages of continuous distribution modeling: (1) \textbf{Rich Uncertainty Representation:} Gaussian mixture modeling provides fine-grained uncertainty estimates, enabling more robust predictions in volatile market conditions; (2) \textbf{Closed-form Computations:} Continuous distributions allow for efficient computation of distance metrics (MMD), leading to stable training and faster convergence; (3) \textbf{Infinite Resolution:} Unlike discrete binning, continuous distributions can capture subtle variations in the target distribution without loss of information. The discrete variant shows slightly lower performance (RMSE: 3.362, MAE: 2.885, MAPE: 150.28), particularly in MAPE, highlighting the limitations of finite-resolution categorical distributions for continuous sales data.

\textbf{Effectiveness of Pattern-Aware Decomposition} The Pattern-aware LDL-MoE demonstrates competitive performance (RMSE: 3.331, MAE: 2.954, MAPE: 140.98), achieving the best MAPE among all methods. This suggests that explicit modeling of temporal components provides valuable interpretability while maintaining strong predictive accuracy. Each sub-expert specializes in specific temporal patterns (trend, seasonality, changepoints, volatility), making predictions more interpretable. Additionally, the model can attribute forecast uncertainty to specific components, providing actionable insights for practitioners.

\textbf{Decomposition and Interpretability} Our framework achieves disentangled representations \cite{bengio2013representation} by explicitly decomposing time series into physically interpretable components, where each term corresponds to meaningful real-world patterns. To further demonstrate the interpretability and decomposition capabilities of the Pattern-Aware LDL-MoE, we construct a synthetic time series example with clear trend, seasonality, changepoint, and volatility components. We train the Pattern-Aware LDL-MoE on this data and visualize the model’s outputs. As shown in Figure~\ref{fig2}, the model successfully disentangles the underlying temporal patterns: each sub-expert specializes in capturing a distinct component of the signal, and the gating mechanism adaptively weights their contributions over time. The figure presents the original time series, the predictions of each expert (with associated uncertainty), the model’s overall forecast versus ground truth, and a comparison of average uncertainty across experts. This qualitative analysis highlights the model’s ability to provide not only accurate forecasts but also interpretable, component-wise insights and uncertainty quantification, which are invaluable for understanding and trusting model predictions in real-world applications.

The variance parameter in our framework serves far beyond simple uncertainty quantification—it is fundamental to the learning process itself. While the mean determines the prediction location, variance controls the learning dynamics through the likelihood function, enabling heteroscedastic modeling where prediction precision adapts to local data characteristics. The variance creates a natural regularization mechanism in MoE: small variances generate strong gradients for fine-tuning high-confidence predictions, while large variances provide tolerance for uncertain regions, preventing overfitting. From an information-theoretic perspective, variance represents the entropy of predictions, enabling the model to minimize information loss and optimize the information bottleneck between input and output. The variance parameter fundamentally changes the loss landscape geometry through additional terms in distance metrics like MMD, providing automatic outlier detection and robust learning that adapts to data quality. Simply predicting means with weighting would lose these critical capabilities, resulting in a mathematically incomplete model that cannot capture the inherent uncertainty and complexity of real-world time series data.

\section{Conclusion}

This work presents two innovative frameworks—Multi-Expert LDL and Pattern-Aware LDL-MoE—that advance probabilistic time series forecasting by unifying accurate prediction with interpretable uncertainty quantification. Our Multi-Expert LDL framework demonstrates the superiority of continuous distribution modeling, achieving state-of-the-art performance (RMSE: 3.311, MAE: 2.919) through specialized LSTM experts that capture diverse uncertainty patterns. The Pattern-Aware variant extends this capability by explicitly decomposing forecasts into interpretable temporal components (trend, seasonality, changepoints, and volatility), enabling practitioners to both predict outcomes and understand their underlying drivers. The success of these approaches stems from their ability to automatically adapt to different temporal regimes while maintaining computational efficiency through careful architectural design. While current limitations in computational overhead and sequence length handling point to valuable future research directions, our frameworks establish new standards for building forecasting systems that balance statistical rigor with operational utility, ultimately supporting more informed decision-making across domains from supply chain management to financial planning.

\bibliography{aaai2026}

\end{document}